\documentclass[runningheads]{llncs}
\usepackage[T1]{fontenc}
\usepackage{graphicx}%
\usepackage{multirow}%
\usepackage{amsmath,amssymb,amsfonts}%
\usepackage{mathrsfs}%
\usepackage[title]{appendix}%
\usepackage{xcolor}%
\usepackage{textcomp}%
\usepackage{manyfoot}%
\usepackage{booktabs}%
\usepackage{algorithm}%
\usepackage{algorithmicx}%
\usepackage{algpseudocode}%
\usepackage[caption=false,font=normalsize,labelfont=sf,textfont=sf]{subfig}
\usepackage{listings}%
\usepackage{verbatim}
\begin{document}
%
\title{Data-Efficient Interactive Multi-Objective Optimization Using ParEGO\thanks{This work has been supported by the Flemish Government under the ‘Onderzoeksprogramma Artificiële Intelligentie (AI) Vlaanderen’ and the ‘Fonds Wetenschappelijk Onderzoek (FWO, grant 12AZF24N)’ programmes.}}
%
%
%

\author{Arash Heidari\inst{1} \and
Sebastian Rojas Gonzalez\inst{1,2} \and
Tom Dhaene\inst{1} \and
Ivo Couckuyt \inst{1}}

\authorrunning{A. Heidari et al.}
%

\institute{Faculty of Engineering and Architecture, Ghent University - imec, Ghent, Belgium
\and Data Science Institute, Hasselt University, Belgium}

\maketitle              
\begin{abstract}
Multi-objective optimization is a widely studied problem in diverse fields, such as engineering and finance, that seeks to identify a set of non-dominated solutions that provide optimal trade-offs among competing objectives. However, the computation of the entire Pareto front can become prohibitively expensive, both in terms of computational resources and time, particularly when dealing with a large number of objectives. In practical applications, decision-makers (DMs) will select a single solution of the Pareto front that aligns with their preferences to be implemented; thus, traditional multi-objective algorithms invest a lot of budget sampling solutions that are not interesting for the DM. In this paper, we propose two novel algorithms that employ Gaussian Processes and advanced discretization methods to efficiently locate the most preferred region of the Pareto front in expensive-to-evaluate problems. Our approach involves interacting with the decision-maker to guide the optimization process towards their preferred trade-offs. Our experimental results demonstrate that our proposed algorithms are effective in finding non-dominated solutions that align with the decision-maker's preferences while maintaining computational efficiency.

\keywords{Multi-Objective Optimization  \and Bayesian Optimization \and Interactive Optimization \and Surrogate Modelling.}
\end{abstract}
\section{Introduction}\label{sec1}
Traditionally, optimization problems focus on a single objective, typically minimizing cost. However, often real-world decision-making requires considering multiple conflicting objectives simultaneously. This gives rise to multi-objective or multi-criteria optimization problems, where performance is assessed using a vector of outputs representing the optimal trade-offs between the conflicting objectives (referred to as the non-dominated or Pareto-optimal set). 

Multi-objective optimization is a well-established field with extensive research \cite{emmerich2018tutorial,li2015many}. Traditional \emph{a posteriori} methods aim to approximate the entire Pareto front first, and then allow the DM to select their preferred solution. Evidently, this is data-inefficient. \emph{A priori} methods on the other hand, utilize a weighted cost function defined before optimization to find a single solution, which requires extensive or complete knowledge of the objective space, which is unrealistic in practical scenarios. 

Our focus in this study is on a potentially more efficient approach, but under-explored in the literature: to solve the problem interactively by iteratively collecting preference information from the decision maker during the optimization process. This interactive approach may lead to the convergence of the most preferred solution(s), without wasting large amounts of computational budget \cite{InteractiveSurvey}. 

The literature on interactive optimization comes from different streams of research; see for instance analytical methods \cite{armbruster2015decision,wallenius2008multiple}, evolutionary Multi-objective Optimization (EMO) algorithms \cite{branke2016using,tomczyk2019decomposition}, and Bayesian multi-objective optimization (BMO) techniques \cite{Astudillo20a,lin2022preference}. Preference-based EMO algorithms have shown efficacy in black-box optimization of complex problems; however, they suffer from data inefficiency, implying the requirement of numerous function evaluation. On the other hand, BMO offers data efficiency for expensive-to-evaluate or time-consuming problems, but incurs an additional modelling cost and error due to the surrogate model. 

Recently, researchers have proposed several variants of BMO interactive methods to guide the search towards the most preferred solution, including scalarization \cite{Branke_PPSN2022_Single}, acquisition functions that capture the DM's preferences and provide theoretical guarantees, but require significant computational resources \cite{Astudillo20a,lin2022preference}, and using reference points as preferences \cite{Taylor2021,Gaudrie2020}. In \cite{Arash_PPSN}, BO is used to identify a desirable solution that is interesting but not necessarily the most preferred solution. This solution is referred to as the knee of the Pareto front, which offers a good trade-off between different objectives.

In this work we introduce two novel scalarization-based BMO approaches to efficiently locate the preferred region of the Pareto front in an interactive and data-efficient manner, relying on Gaussian Processes to build metamodels of the objectives and advanced discretization techniques to explore interesting regions of the search space. Due to space constraints we don't discuss Bayesian Optimization, but please refer to \cite{frazier2018bayesian} for a comprehensive overview.

\section{Proposed Methods}

Throughout this paper, an interaction with the DM involves presenting the extracted Pareto front and soliciting their preference for the most desirable solution so far. The first method employs a Delaunay triangulation technique \cite{gramacy2022triangulation} to partition the input space and subsequently explores the preferred region through DM feedback. The second approach involves sampling and adjusting the scalarization weights to encourage exploration of the preferred region. Both methods consist of an initialization phase and an exploration phase; the initialization phase is based on the well-known ParEGO algorithm \cite{ParEGO}, and it is the same for both approaches. In what follows we first introduce the Initialization phase, followed by the description of both methods proposed. 

\subsection{Initialization Phase}
The initialization phase itself comprises two steps. First, $P_{space}$ candidates are evaluated using a space-filling design such as Halton sampling with the expensive-to-evaluate function $F(\cdot)$. In the second step, the augmented Tchebycheff scalarization function is employed with $P_{init}$ weights sampled uniformly from a $d_{out}$ dimensional probability simplex, $\Delta^{d_{out}}$ (defined in equation \ref{equ-simplex}), where $d_{out}$ is the number of objectives. Since the weights are sampled i.i.d. uniformly, the resulting evaluations of the proposed candidates are expected to be on or close to the Pareto front with high diversity \cite{ParEGO}. The non-dominated solutions obtained after executing the initialization phase are subsequently presented to the DM to select their desired trade-off (i.e., their preferred solution thus far). Algorithm \ref{alg-initialization} outlines the specific steps involved in the initialization phase. It should be noted that in Step 4 of the algorithm, a small value for $\rho$ (e.g., 0.05) is used.

\begin{equation}
    \Delta^{d_{out}} = \bigg\{ \mathbf{w} \in \mathbb{R}^{d_{out}} \ \bigg| \ \sum_{i=1}^{d_{out}} w_i = 1, \ w_i \geq 0 \ \text{ for all } i \bigg\}
\label{equ-simplex}
\end{equation}

\begin{algorithm}
\caption{Initialization Phase}\label{alg-initialization}
\begin{algorithmic}[1]
\algrenewcommand\algorithmicrequire{\textbf{Input}}
\Require $d_{out}$: number of objectives
\Require $\Delta ^ {n}$: $n$-dimensional simplex 
\Require $\alpha(\cdot)$: EI Acquisition function \Comment{Eq. \ref{eq:EI}}
\State $D = \{x_i, F(x_i)\ | x_i \sim Halton Sampling\}_{i=1}^{P_{space}}$ 
\State $\mathbf{W_{init}} = \{\mathbf{w_j} | \mathbf{w_j} \sim \Delta ^ {d_{out}}\}_{j=1}^{P_{init}}$  
\For{$\mathbf{w}$ in $\mathbf{W_{init}}$}
    \State $U = max_i[w_i \cdot F_i] + \rho \sum_{j=1}^{k} [w_i \cdot F_i]$ \Comment{Tchebycheff scalarization}
    \State $\Tilde{M} \gets$ Surrogate model for $U$ \Comment{Employing GP}
    \State $c = \mathop{arg max}_{x} \alpha(x; \Tilde{M}, D)$
    \State $D = D \cup \{c, F(c)\}$
\EndFor
\State Interact with the DM for feedback
\State $P \gets$ The preferred solution on the extracted Pareto front
\State $\mathbf{w_p} \gets$ The corresponding weight for $P$
\State $\mathbf{x_P} \gets$ The corresponding input for $P$
\end{algorithmic}
\end{algorithm}

Algorithm \ref{alg-initialization} incorporates the Expected Improvement (EI) acquisition function, which is defined as follows:

\begin{equation}
    EI(\mathbf{x_*}) = (f_{min}-\mu(\mathbf{x_*}))\Phi\left(\frac{f_{min}-\mu(\mathbf{x_*})}{s(\mathbf{x_*)}}\right)+s(\mathbf{x_*})\phi\left(\frac{f_{min}-\mu(\mathbf{x_*})}{s(\mathbf{x_*})}\right)
\label{eq:EI}
\end{equation}

\noindent Here, $\mu(\mathbf{x_*})$ and $s(\mathbf{x_*})$ represent the prediction and uncertainty of the surrogate model at the new point $\mathbf{x_*}$, respectively. $f_{min}$ denotes the minimum value extracted thus far, and $\Phi$ and $\phi$ represent the cumulative distribution function and the probability density function, respectively.

\subsection{Triangulation-based Preferred Region Exploration (TRIPE)}

\begin{figure}[h]
  \centering
  \includegraphics[width=0.9\linewidth]{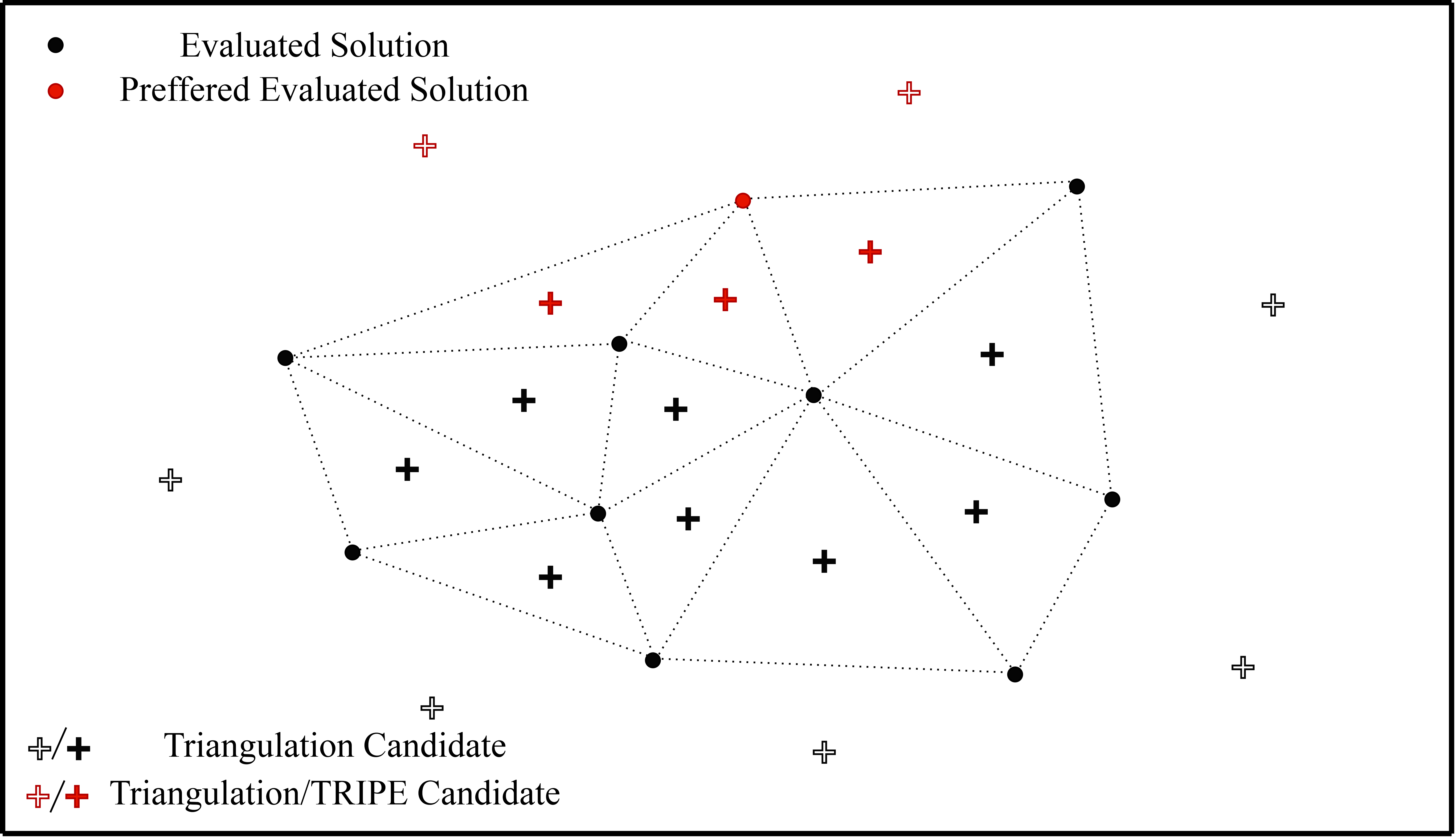}
  \caption{Triangulation in a 2-dimensional input space. Circles represent the current solutions, with the red circle indicating the solution chosen by the DM. Crosses represent the candidates generated by triangulation. The red-colored crosses, which are neighbors of the preferred solution, will be evaluated in the next iteration.}
  \label{fig-triangulation}
\end{figure}

The initialization phase yields a set of initial candidates and weights, which are subsequently employed in the optimization process. TRIPE employs a sophisticated method to discretize the input space and explore the region near the preferred solution using Delaunay triangulation candidates \cite{Lee1980}. A Delaunay triangulation is a set of line segments that connects nearby points in a manner that maximizes the angles while ensuring that no point lies inside the circumcircle of those points. This triangulation method subdivides the interior of the smallest convex set, which contains all points, known as the convex hull. The center of each triangle serves as a potential candidate for the next evaluation, referred to as the \emph{interior candidates} (solid crosses in Fig. \ref{fig-triangulation}). However, as found in \cite{gramacy2022triangulation}, limiting the exploration to only interior candidates constrains the exploration unless all corner points are included in the initial design. Therefore, TRIPE generates additional candidates outside any triangle as suggested in \cite{gramacy2022triangulation}, referred to as \emph{fringe candidates} (hollow crosses in Fig. \ref{fig-triangulation}). Interested reader are encouraged to refer to \cite{gramacy2022triangulation} for the calculation of the interior and fringe candidates. Note that in the work by Gramacy et al. \cite{gramacy2022triangulation}, this triangulation method was employed for single-objective Bayesian optimization; here, we have adapted this approach to explore the preferred regions of the input space in a multi-objective scenario. Specifically, we select the direct neighbours of the selected preferred solution for expensive evaluation, as illustrated in Fig.~\ref{fig-triangulation}. Algorithm \ref{alg-method1} presents an overview of the TRIPE exploration phase.

\begin{algorithm}
\caption{TRIPE Exploration Phase}\label{alg-method1}
\begin{algorithmic}[1]
\algrenewcommand\algorithmicrequire{\textbf{Input}}
\Require $P$: Preferred solution
\Require $\mathbf{x_P}$: Corresponding input of the preferred solution
\While{$Budget$ left}
    \State Delaunay triangulation of the Input Space
    \State $C_I \gets$ The interior candidates neighbouring $\mathbf{x_P}$
    \State $C_F \gets$ The fringe candidates neighbouring $\mathbf{x_P}$
    \State Evaluate $C_I \cup C_F$
    \State Interact with the DM about her preference.
    \State Update $P$ and $\mathbf{X_P}$
\EndWhile
\end{algorithmic}
\end{algorithm}

As pointed out in \cite{gramacy2022triangulation}, the Delaunay triangulation and convex hull algorithms, which rely on existing libraries, can be computationally demanding for large candidate sets (e.g., thousands), as required for high-dimensional problems. Thus, TRIPE is best suited for low-dimensional problems, typically with no more than five input dimensions.

\subsection{Weight Adjustment for Preferred Region Exploration (WAPE)}

The exploration phase of WAPE follows a similar procedure as the initialization phase. During the initialization phase, the weights for the scalarized function are uniformly sampled from a $d_{out}$-dimensional probability simplex. However, in the exploration phase of WAPE, the $N$ newly generated weights are constrained to be in the \textit{proximity} of the preferred weight $\mathbf{w_p}$. The generation of a new weight utilizes Equation \ref{equ-weight-modif}, where $\eta$ denotes the \textit{explore} parameter that determines the extent of the neighbouring region around the preferred solution to be explored. Smaller values of $\eta$ result in new candidates being located closer to the preferred solution, while larger values allow for a broader search in the objective space. Additionally, $N$ controls the number of newly generated weights in the vicinity of the preferred weight. Smaller values of $N$ encourage more interaction with the DM, while larger values aim to minimize interactions, particularly when DM interaction is resource-intensive. Algorithm \ref{alg-method2} provides an overview of the WAPE approach.

\begin{equation}
    \mathbf{w_{new}} = \frac{\mathbf{w_P} * \mathbf{\theta}}{\sum_i{w_{P,i} * \theta_i}},
\label{equ-weight-modif}
\end{equation}
\begin{displaymath}
  \mathbf{\theta} \sim U(1 - \eta, 1 + \eta) ^ {d_{out}}
\end{displaymath}

\begin{algorithm}
\caption{WAPE Exploration Phase}\label{alg-method2}
\begin{algorithmic}[1]
\algrenewcommand\algorithmicrequire{\textbf{Input}}
\Require $N$: number of new candidates
\Require $w_{P}$: Corresponding weight for the preferred solution
\Require $D$: Evaluated design of experiment
\Require $\alpha(\cdot)$: EI Acquisition function \Comment{Eq. \ref{eq:EI}}
\While{$Budget$ left}
    \State $\mathbf{W_{new}} = \{\mathbf{w_i} | \mathbf{w_i} = \frac{\mathbf{w_P} * \mathbf{\theta}}{\sum_i{w_{P,i} * \theta_i}} \} _{i=1}^ N$ \Comment{Generating $N$ new weights using Eq. \ref{equ-weight-modif}}
    \For{$\mathbf{w}$ in $\mathbf{W_{new}}$}
        \State $U = max_i[w_i \cdot F_i] + \rho \sum_{j=1}^{k} [w_i \cdot F_i]$ \Comment{Tchebycheff scalarization}
        \State $\Tilde{M} \gets$ Surrogate model for $U$ \Comment{Emplying GP}
        \State $c = \mathop{arg max}_{x} \alpha(x; \Tilde{M}, D)$
    \State $D = D \cup \{c, F(c)\}$
    \EndFor
    \State Interact with the DM about her preference
    \State Update $\mathbf{w_P}$
\EndWhile
\end{algorithmic}
\end{algorithm}

\section{Results}
Standard performance measures designed for evaluating the quality of an entire Pareto front approximation obtained by an optimizer, such as hypervolume or inverted generational distance, are not suitable for assessing preference-based multi-objective optimizers, as noted in \cite{afsar2021assessing}. In the absence of a DM, it is difficult to ascertain whether the algorithm is converging towards the most preferred solution. Hence, this study employs a recently proposed metric in \cite{Branke_PPSN2022_Single}, which measures the \emph{Opportunity Cost} (OC) between a pre-defined ground truth and any other evaluated solution (see Equation \ref{equ-oc}). Here, $x^*$ represents the ground truth in the input space, and $x$ can be any evaluated solution up to that point. The preference of the DM is selected as the input with the lowest OC.

\begin{equation}
    OC = U(F(x^*)) - U(F(x))
\label{equ-oc}
\end{equation}

To determine the ground truth solution $x^*$ in Equation \ref{equ-oc}, we first apply an EMO algorithm (NSGA-II) to find the entire Pareto front of the problem. We then use the preferred weights, randomly selected by sampling from a $d_{out}$ dimensional probability simplex, to define a scalarization function and select the solution in the Pareto front that minimizes this function as the ground truth $x^*$. Our proposed algorithms are evaluated using $P_{space} = 10 \times d_{in}$ and $P_{init} = 10 \times d_{out}$, where $d_{in}$ and $d_{out}$ represent the input and output dimensions, respectively. We also consider the cost of interaction with the decision maker, $Cost_{DM}$, which is equivalent to the cost of one expensive evaluation. Each experiment is repeated 10 times, and we report the median as well as the $20^{th}$ and $80^{th}$ percentiles. 

\begin{figure*}[h]
\centering
\subfloat[]{\includegraphics[width=1.5in]{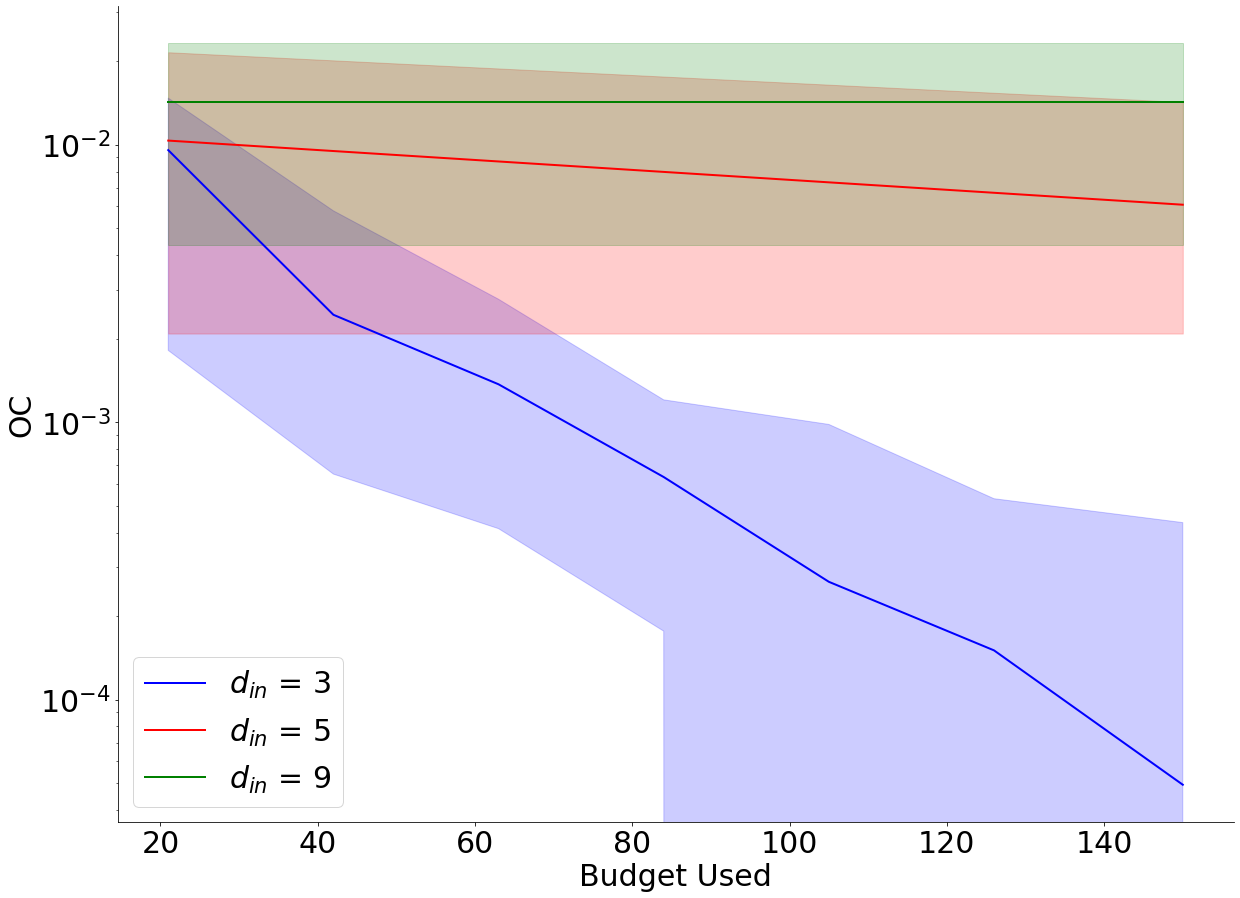}%
\label{effect-input-tc}}
\hfil
\subfloat[]{\includegraphics[width=1.5in]{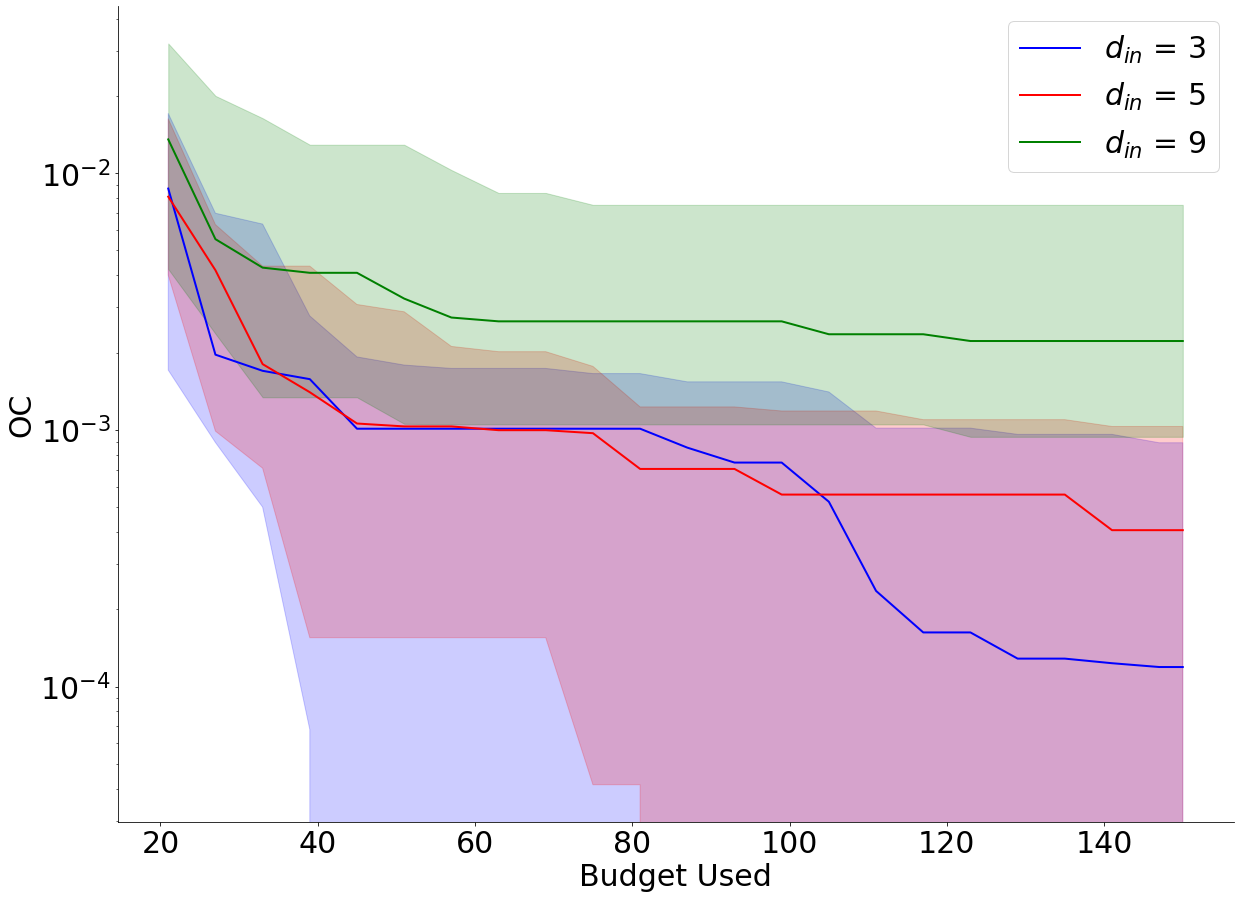}%
\label{effect-input-wm}}
\hfil
\subfloat[]{\includegraphics[width=1.5in]{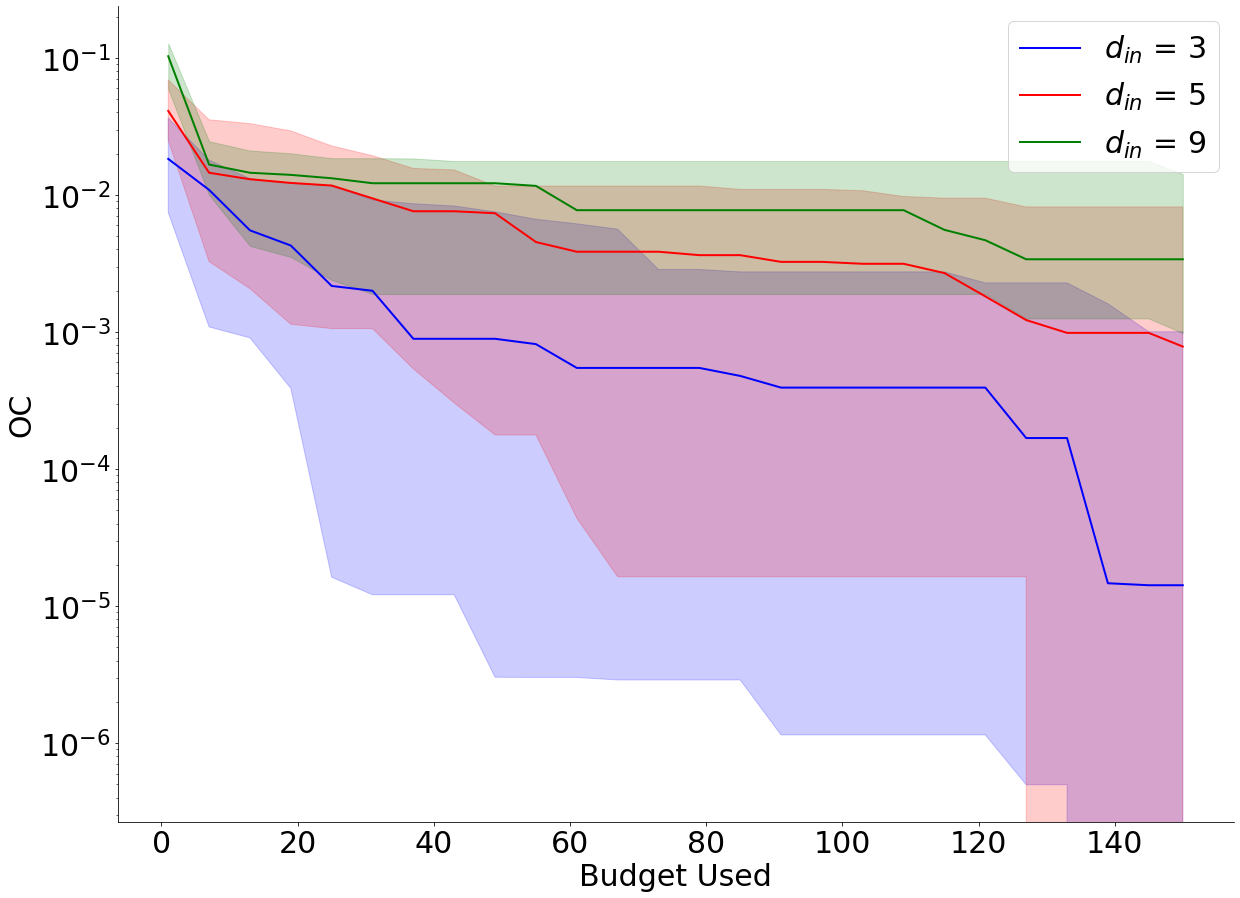}%
\label{effect-input-oup}}

\caption{The OC during optimization for (a) TRIPE and (b) WAPE (c) benchmark \cite{Hakanen2017}.}
\label{fig-effect-input}
\end{figure*}

We investigate the performance of the proposed methods against a competing benchmark from the literature \cite{Hakanen2017} on the DTLZ2 problem with two objectives and 3, 5 and 9 input dimensions. The hyperparameters of WAPE are set to $N=5$ and $\eta=0.05$. The results presented in Figure \ref{fig-effect-input} demonstrate that the WAPE approach exhibits superior performance to TRIPE and the benchmark from the literature. Notably, we observe that the performance trend of the objective convergence for TRIPE does not improve with increasing computational budget when the input dimension is 5 or greater.

\section{Conclusion}
We have proposed two new scalarization-based Bayesian algorithms for interactive optimization of multi-objective problems with limited evaluation budgets. The TRIPE method uses a triangulation-based strategy that considers the neighbours of a given preferred solution in input space. The algorithm is hyperparameter-free and utilizes state-of-the-art computational libraries for triangulation and convex hull calculations, but is limited when scaling to more than 5 input dimensions. On the other hand, WAPE can handle more complex problems with higher input dimensions but is computationally more expensive as it uses a secondary surrogate model and acquisition function. Our preliminary results show that both proposed methods are competitive against a benchmark from the literature, and for further work we intend to evaluate more difficult test problems.


%
%
%
\bibliographystyle{splncs04}
%
\bibliography{bib}

\end{document}